\documentclass[letterpaper, 10 pt, conference]{ieeeconf}  

\IEEEoverridecommandlockouts                              

\usepackage{graphics} 
\usepackage{epsfig} 
\usepackage{amsmath} 
\usepackage{amsfonts}
\usepackage{amssymb}  
\usepackage{mathtools, mathdots}
\usepackage{algorithmic,algorithm}
\usepackage{subcaption}
\usepackage{multirow}
\usepackage{booktabs} 
\usepackage{textcomp}
\usepackage[usenames,dvipsnames,svgnames,table, xcdraw]{xcolor}
\usepackage{hyperref}
\hypersetup{
    colorlinks=true,
    linkcolor=blue,
    filecolor=magenta,
    urlcolor=blue,
}
\usepackage{todonotes}
\definecolor{objects}{rgb}{0.09, 0.75, 0.82}
\definecolor{table}{rgb}{1.0, 0.5, 0.01}
\definecolor{cabinet}{rgb}{0.122, 0.467, 0.706}
\definecolor{wall}{rgb}{0.682, 0.780, 0.902}
\definecolor{clothes}{rgb}{0.807, 0.427, 0.745}
\definecolor{mirror}{rgb}{0.859, 0.859, 0.5449}
\definecolor{picture}{rgb}{0.839, 0.153, 0.157}
\definecolor{lighting}{rgb}{0.710, 0.812, 0.416}

\definecolor{ceiling}{rgb}{0.612, 0.613, 0.875}
\definecolor{appliances}{rgb}{0.906, 0.796, 0.588}
\definecolor{sink}{rgb}{0.518, 0.235, 0.224}
\definecolor{stool}{rgb}{0.388, 0.475, 0.243}
\definecolor{plant}{rgb}{0.549, 0.635, 0.314}
\definecolor{counter}{rgb}{0.647, 0.318, 0.588}
\title{\LARGE \bf
Active Semantic Mapping and Pose Graph Spectral Analysis for Robot Exploration
}

\author{Rongge Zhang,
        Haechan Mark Bong,
        Giovanni Beltrame
        \thanks{The authors are with the Department of Computer Engineering and Software Engineering, Polytechnique Montréal, Montréal, Canada. {\tt\small \{rongge.zhang, haechan.bong, giovanni.beltrame\}@polymtl.ca}}
        \thanks{This work was supported by the National Research Council Canada (NRC).}
}

\begin{document}

\maketitle
\thispagestyle{empty}
\pagestyle{empty}

\begin{abstract}
  Exploration in unknown and unstructured environments is a pivotal requirement
  for robotic applications. A robot's exploration behavior can be inherently
  affected by the performance of its Simultaneous Localization and Mapping
  (SLAM) subsystem, although SLAM and exploration are generally studied
  separately. In this paper, we formulate exploration as an active mapping
  problem and extend it with semantic information. We introduce a novel active
  metric-semantic SLAM approach, leveraging recent research advances in
  information theory and spectral graph theory: we combine semantic mutual
  information and the connectivity metrics of the underlying pose graph of the
  SLAM subsystem. We use the resulting utility function to evaluate different
  trajectories to select the most favorable strategy during exploration.
  Exploration and SLAM metrics are analyzed in experiments. Running our
  algorithm on the Habitat dataset, we show that, while
  maintaining efficiency close to the state-of-the-art exploration methods, our
  approach effectively increases the performance of metric-semantic SLAM with a
  21\% reduction in average map error and a 9\% improvement in average semantic classification
  accuracy.

\end{abstract}

\section{Introduction}
Autonomous exploration and scene understanding constitute pivotal capabilities for advancing robotic systems. Conventionally, exploration relies on geometric maps to facilitate navigation and obstacle avoidance. However, augmenting a robot's autonomy necessitates a comprehensive understanding of the environment's semantic information. Metric-semantic Simultaneous Localization and Mapping (SLAM) emerges as a key solution, offering a dual perspective of geometric and semantic environmental perception. This is invaluable for wide applications, from indoor service robots to large-scale outdoor exploration. However, current approaches often employ passive robot motion control during SLAM, limiting the potential for dynamic exploration.

In previous literature on robotic exploration, the information-theoretic approaches are promising solutions, showing the ability to facilitate rapid exploration and minimize map uncertainty. Nevertheless, these approaches often segregate the active mapping problem from localization, i.e., assuming that localization errors have been resolved. This assumption can introduce constraints in practical scenarios, such as compromised path planning due to SLAM errors and incorrect object labeling in semantic maps, causing the robot to pick up wrong objects.

\begin{figure}[t]
\centerline{\includegraphics[width=1.0\linewidth]{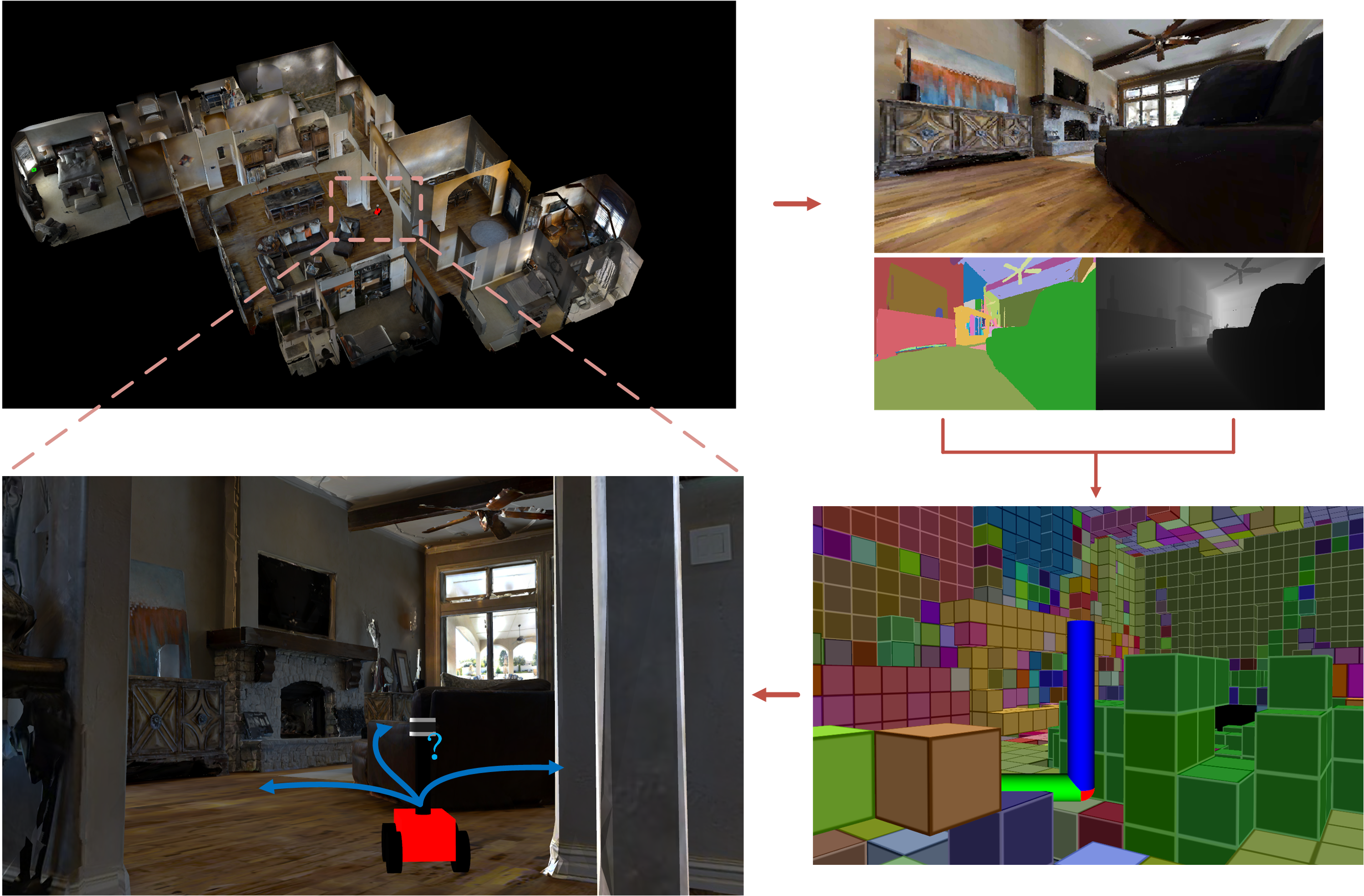}}
\caption{Representation of our problem: a robot explores a Habitat environment,
  building a 3D semantic map using depth sensors and segmentation, and using it for navigation.}
\label{fig1}
\end{figure}

Motivated by these gaps, this paper focuses on active metric-semantic SLAM. Our approach extends existing information-theoretic methods for binary~\cite{doi:10.1177/0278364920921941} or multi-class~\cite{asgharivaskasi2023semantic} active mapping by incorporating localization uncertainty into exploration. We maintain a pose graph and utilize a novel quantitative metric of pose graph uncertainty, inspired by recent advances in spectral graph theory. Analyzing the semantic mutual information and underlying pose graph topology allows us to evaluate the maximum gain of different potential trajectories, thus promising higher-quality localization and semantic mapping while maintaining efficient exploration (see Fig. \ref{fig1} as an illustration). In summary, the main contributions of this paper are:
\begin{enumerate}
    \item A novel approach to active SLAM with online decision-making for autonomous robot exploration;
    \item A real-time metric-semantic mapping system for active robot perception and scene understanding;
    \item A hybrid representation integrating semantic mutual information gain and localization uncertainty to guide the robot's action policy;
\end{enumerate}
We demonstrate the performance of our method through extensive experiments in photorealistic environments, our implementation is publicly available at: https://github.com/BohemianRhapsodyz/semantic\_exploration.

\section{Related Work}

\subsection{Semantic Mapping}
Semantic mapping aims to construct semantically annotated 3D maps during robotic operations. Initial approaches focused on embedding semantic labels into the geometric spatial representation of the map such as voxel, surfel, or mesh, updating them via Bayesian probabilistic fusion upon receiving new observations, typically by SemanticFusion~\cite{7989538} and Voxblox++ \cite{grinvald2019volumetric}.
To achieve higher-level scene understanding, hybrid representations such as scene graphs are introduced, pioneering works include Kimera~\cite{Rosinol20icra-Kimera} and Hydra~\cite{Rosinol20rss-dynamicSceneGraphs}. Recent works like Concept-Graph~\cite{conceptgraphs}, which implements a multimodal correspondence between language and vision based on Contrastive Language-Image Pre-training (CLIP) \cite{radford2021learning}, can be used for text querying and language interaction navigation. Moreover, neural representations have emerged as a promising frontier in semantic mapping, although they often come with computational overhead and real-time constraints. Examples include Neural Radiance Fields (NeRF)~\cite{rosinol2023nerf} and Gaussian-Splatting-based mapping methods~\cite{keetha2023splatam}.

\subsection{Exploration}
Exploration strategies in robotics have significantly evolved since the seminal concept of frontier exploration, different exploration methods have been proposed, ranging from frontier selection to optimization of travel costs. For the application in recent years, GBPlanner~\cite{dang2020graph} was successfully used in the DARPA Subterranean Challenge. TARE planner~\cite{cao2023representation} proposed a hierarchical exploration strategy directly on point cloud maps and has been extended to multi-robot exploration.
Among these different methods, information-theoretic exploration aims to achieve the balance between unexplored space and robot state uncertainty. Approaches like Cauchy-Schwarz quadratic mutual information (CSQMI)~\cite{7139865} and fast Shannon mutual information (FSMI)~\cite{doi:10.1177/0278364920921941} have achieved fast exploration on the binary-class map. Semantic Shannon mutual information (SSMI)~\cite{asgharivaskasi2023semantic} followed up by developing a closed-form expression for multi-class mutual information and successfully showing its capabilities for exploration with semantic maps. A common limitation across these methodologies is the assumption of perfect localization, which is impractical in reality. Recent work~\cite{chen2023train} evaluates the frontier gain and considers localization uncertainty. Through ablation experiments, it elucidates the importance of localization uncertainty in influencing exploration performance. However, it stops short of addressing the navigation challenges this uncertainty introduces.

\subsection{Active SLAM}
Active SLAM is usually considered to be the problem of reducing SLAM localization and mapping uncertainty by actively controlling the motion of the robot (or sensors attached to the robot) and is often decomposed into three phases: goal determination, utility computation, and action selection~\cite{placed2023survey}. Carlone et al.~\cite{carlone2014active} propose a utility function for discounting the information gain of an action based on the probability of having good localization and intentionally guiding the robot to revisit previously explored environments to increase potential loop closure constraints. However, their approach is limited to particle-filtered SLAM. Branching off information theory, the theory of optimal experimental design (TOED) is introduced to evaluate the uncertainty of potential trajectories, with Carrillo et al.~ \cite{6224890} showing that D-optimality, which guarantees monotonicity during exploration, is an ideal criterion for TOED and proposing Shannon-Rényi entropy~\cite{7139224} to evaluate varying magnitudes of map and localization uncertainty. Placed et al.~\cite{placed2021fast} present a general relationship between optimality criteria and connectivity indices for active graph-SLAM and use it for fast exploration. However, the map points are marginalized during the construction of the pose graph, the expected gain is calculated as a simple unexplored cell ratio, and the uncertainty of the map is not adequately taken into account. In recent work, Tao et al.~\cite{tao20243d} propose a 3D active SLAM system for drones by using semantic information for detecting loop closure and estimating the relative pose transformations. Distinguishing from their work, this paper uses semantics for mutual information computation and focuses on the utility function in active SLAM.

\subsection{Spectral Graph Theory}
Optimal design criteria, such as D-optimality, offer a scalar representation of the information matrix of the SLAM estimation, effectively quantifying the uncertainty. While these metrics have significantly advanced active SLAM research, evaluating them involves computing a large and dense covariance matrix, which is still expensive, limiting its usage for real-time decision-making. Notably, it has been shown that there is a close relationship between the optimality and topology structure of the underlying pose graph~\cite{placed2022general}. Specifically, maximizing optimality equates to maximizing the algebraic connectivity of the underlying graph, which can be determined by analyzing the Laplacian spectrum of the topology, thus avoiding the costly computational burden associated with calculating the entire posterior covariance matrix. This has spurred many advancements in active SLAM~\cite{placed2022explorb}. Interestingly, beyond active SLAM, the study of spectral graph theory has also been widely applied in other SLAM fields such as rotation averaging and translation estimation~\cite{tian2023spectral}, pose graph sparsification~\cite{doherty2022spectral} and collaborative SLAM~\cite{lajoie2023swarm}.

\section{Preliminaries}
\subsection{Occupancy Grid Mapping and Mutual Information}
As is standard in information-theoretic exploration, we model the environment in occupancy grid. The map $m$ consists of $n$ grids that can be modeled as $m = \{ {m_1},{m_2}, \cdots ,{m_n}\} $, where ${m_i} \in k$ indicates $k$ categories. For the binary mapping, $k \in \{ 0,1\} $ for free and occupied cells, respectively. The observation is from a range measurement sensor such as LiDAR or RGB-D camera equipped on the robot. Assume we have an observation ${Z_t}$ and robot pose ${X_t}$ at time $t$, the probability density function (PDF) of map ${p_t}(m)$ can be updated using Bayesian rule by:
\begin{equation} \label{eq1}
{p_t}(m) \propto p({Z_t}\left| {m,{X_t}} \right.){p_{t - 1}}(m).
\end{equation}
Mutual information (MI) is introduced and devoted to reducing mapping uncertainty (entropy). Given a new observation, minimizing the map's conditional entropy is equal to maximizing the MI between $m$ and ${Z_t}$:
\begin{equation} \label{eq2}
\begin{split}
& I\left(m ; Z_t=z_t \mid z_{1: t-1}, x_{1: t-1}\right) =\\
&H\left(m \mid z_{1: t-1}, x_{1: t-1}\right)-H\left(m \mid Z_t=z_t, z_{1: t-1}, x_{1: t-1}\right).
\end{split}
\end{equation}

For a beam-based sensor with finite range $\Theta$ beams observation, the MI between the map and the measurements is the sum of the MI for every beam $\theta  \in \Theta $ in a sensor scan at time $t$:
\begin{equation} \label{eq3}
\begin{aligned}
I(m;{Z_{t,\Theta }}&\left|{{Z_{1:t - 1,\Theta }}} \right.) = \int\limits_{\theta  \in \Theta } {I(m,{Z_{t,\theta }})d\theta } \\
&= \int_{\theta  \in \Theta } {\textstyle\sum\limits_{m \in k} {p(m,{Z_{t,\theta }}\left| {{Z_{1:t - 1,\theta }}} \right.)} } \\
&\times \log \frac{{p(m,{Z_{t,\theta }}\left| {{Z_{1:t - 1,\theta }}} \right.)}}{{p(m\left| {{Z_{1:t - 1,\theta }}} \right.)p({Z_{t,\theta }}\left| {{Z_{1:t - 1,\theta }}} \right.)}}d\theta.     
\end{aligned}
\end{equation}

Thus, by accumulating the MI over the independent grids and counting the overall MI, we can evaluate every potential trajectory and select the one that maximizes the information gain for robotic exploration and active mapping. 

\subsection{Graph-based SLAM and Optimality Criteria}
The SLAM estimation problem can be solved with graph methods, where the node and edge represent the pose and relative motion, respectively. The relative estimation is perturbed with noise under the Gaussian distribution assumption. In this paper, we focus on pose graph optimization, i.e. the landmark or map points are marginalized (the mapping uncertainty will still be considered in the semantic mapping part of this work). In short, the pose graph optimization problem can be treated as a least-square solving problem:
\begin{equation} \label{eq4}
\mathop {\min }\limits_x f(x) = \mathop {\min }\limits_x \frac{1}{2}\textstyle\sum\limits_{i,j \in \varepsilon } {e_{ij}^T(x)\Sigma _{ij}^{ - 1}{e_{ij}}(x)}.
\end{equation}
where $x$ is the state (robot pose), $\varepsilon$ is the set of all edges, ${{e_{ij}}(x)}$ is the error to be minimized and ${\Sigma _{ij}}$ is the covariance matrix. 
Several optimality criteria can be applied to SLAM according to the theory of optimal experimental design. A-optimality, T-optimality, E-optimality, and D-optimality~\cite{6224890} are mostly used criteria that optimize different properties of the covariance matrix in pose graph optimization. 
Among them, only D-optimality captures the global variance uncertainty and keeps the monotonicity~\cite{6224890}, the representation of D-optimality is: 
\begin{equation} \label{eq5}
D \text- opt = \exp (\frac{1}{n}\textstyle\sum\limits_{i = 1}^n {\log{\lambda_i}} )
\end{equation}
where $\lambda_i$ is the eigenvalue of $\Sigma _{ij}$.
\subsection{Optimality Criteria On the Spectra}
Recent research also investigates the optimality of the topological structure i.e. Laplacian matrix of the underlying pose graph. We start by representing the pose graph as a weighted weakly connected directed graph ${\cal G} = ({\cal V},{\cal E},\omega )$, where $\cal V$ and $\cal E$ are vertices and edges of the graph and $\omega$ are the weights for every edge in the graph. The connection relationship between every vertex and edge is defined by the incidence matrix $A = [{a_1},{a_2},...,{a_m}] \in {\{  - 1,0,1\} ^{n \times m}}$. Then the Laplacian matrix corresponding to the directed graph $\cal G$ is defined as: 
\begin{equation} \label{eq6}
L = ADiag\{ {\omega _1},{\omega _2},...,{\omega _m}\} {A^T}.
\end{equation}
The Laplacian matrix has many important properties, which can be derived from the graph connectivity from the analysis of the Laplacian spectrum. Consider that the eigenvalue of $L$ is $\lambda _L^i = (\lambda _L^1,\lambda _L^2,...,\lambda _L^m)$, the second small eigenvalue in $L$, as known as the algebraic connectivity, is a key quantity metric as the inverse of the algebraic connectivity bounds the Cremér-Rao lower bound on the expected mean squared error for averaging \cite{chen2021cramer}. Similarly, the optimality of the Laplacian matrix can also be defined using the number of spanning trees \cite{placed2022general}. Given the number of spanning trees of $\cal G$:
\begin{equation} \label{eq7}
S({\cal G}) = cof(L) = \frac{1}{n}\prod\limits_{i = 2}^m {\lambda _L^i},
\end{equation}
where $cof$ refers to the cofactor of $L$. 
Then the D-optimality of a Laplacian matrix is defined as:
\begin{equation} \label{eq8}
D \text- opt(L) = {(nS({\cal G}))^{\frac{1}{n}}},
\end{equation}
where $n$ is the number of nodes.

\section{Active SLAM Based on Semantic Mutual Information and Pose Graph Spectrum}
\subsection{Semantic Mutual Information}
In Section III-A, we have the methodology for computing mutual information within the occupancy grid map, which enables the robot to explore the trajectory that minimizes map uncertainty. While accounting for the robot kinematics and sensor errors on map entropy, equation~\eqref{eq3} still falls short in active semantic mapping due to its limited capacity to handle semantic segmentation uncertainty. Deriving the representation of semantic mutual information is beneficial for improving the accuracy of the semantic map. For a map $m$ with ${m_i}$ where ${m_i} \in c,c \in [0,C]$ which means we have $C$ different classes of segmentation (0 for free space), the PDF of $m_i$ belongs to one specific class $c$ is:
\begin{equation} \label{eq9}
\begin{split}
p(m_i = c) &= {logit^{-1}}(y) = \frac{{e_{c + 1}^T\exp ({y_i})}}{{{1^T}\exp ({y_i})}} \\
e_{c + 1}^T &= \left\{ \begin{array}{l}
1\;\;\;e_{c + 1}^{T[k]} = 1\\
0\;\;\;other.
\end{array} \right.
\end{split}
\end{equation}
Here we use $logit^{-1}$ to represent the inverse transform of log-odds, as the probability of $y$ over $c$ is expressed in a vector of log-odds:
\begin{equation} \label{eq10}
{y_i} = [\log \frac{{p({m_i} = 0)}}{{p({m_i} = 0)}}, \cdots ,\log \frac{{p({m_i} = C)}}{{p({m_i} = 0)}}].
\end{equation}
The probability $y$ is updated by:
\begin{equation} \label{eq11}
{y_{t + 1,i}} = {y_{t,i}} + \sum ({\beta _i}  - {y_{0,i}}),
\end{equation}
where ${\beta _i}$ is the inverse observation model obtained by Bresenham's line algorithm \cite{5388473}.
To compute the semantic mutual information, instead of accumulating the MI in every grid which may be hit many times and considering that the ray from the sensor can intersect in multiple measurements, we follow the method used in \cite{asgharivaskasi2021active} to select a subset of measurements in which there are no overlapping between rays, this leads to a low bound of the semantic mutual information. For a calculation time horizon $T$, the MI can be approximated as:
\begin{equation} \label{eq12}
\begin{split}
&I(m;{Z_{t:t + T,\Theta }}\left| {{Z_{1:t - 1,\Theta }}} \right.) \ge \int\limits_{\theta  \in \Theta } {I(m,{{\underline Z }_{t:t + T,\theta }})d\theta } \\
 &= \sum\limits_{\tau  = t}^{t + T} {\sum\limits_{\theta = 1}^R {\sum\limits_{i \in {I_\theta }(l\max )} {I(m_i;{Z_{\tau,\theta }}\left| {{Z_{1:t - 1,\theta }}} \right.)} } }. 
\end{split}
\end{equation}
where ${{{\underline Z }_{t:t + T,\theta }}}$ are the selected subset of measurements, $R$ is the max indices of the ray, ${{I_\theta }}$ is the set of map cell indices along the ray and ${l\max }$ is the max sensing range.
Using the expression in equation~\eqref{eq3} and substituting~\eqref{eq9} into~\eqref{eq12}, we have:
\begin{equation} \label{eq13}
\begin{split}
\int\limits_{\theta  \in \Theta }& {I(m,{{\underline Z }_{t + 1:t + T,\theta }})d\theta }  = \\
&\sum\limits_{\tau  = t}^{t + T} {\sum\limits_{\theta  = 1}^R {\sum\limits_{c = 1}^C {\int\limits_0^{l\max } {(p({Z_{t,\theta }} = (l,c)\left| {{Z_{1:t - 1}}} \right.)} } } } \\
&\sum\limits_{i \in {I_\theta }(l\max )} {h{({\beta _i}}(l,c)) - {y_{0,i}},{y_{t,i}}} )dl,
\end{split}
\end{equation}
where 
\begin{equation} \label{eq14}
h(x,y) = \log (\frac{{{1^T}\exp (y)}}{{{1^T}\exp (x + y)}}) + {y^T}logit^{- 1}(x + y).
\end{equation}
To compute the first element in equation~\eqref{eq13}, let $d$ denote the distance traveled by the ray within cell $m_i$, and ${i^*}$ denote the index of the cell hit by ${{Z_{t,\theta }}}$, the first item is calculated by:
\begin{equation} \label{eq15}
p({Z_{t,\theta }} = (l,c)\left| {{Z_{1:t - 1}}} \right.) = \frac{{{p_t}({m_{{i^*}}} = c)}}{{d({i^*})}}\prod\limits_{i \in {I_\theta }(l)\backslash {i^*}} {{p_t}({m_i} = 0)}.
\end{equation}
We utilize semantic mutual information as defined in~\eqref{eq13} instead of relying on binary occupancy probabilities. This approach allows us to evaluate the confidence in the semantic classification of each voxel in the multi-class mapping. A voxel's category probability nearing 1 suggests a higher likelihood of belonging to a specific semantic category, which is also reflected in the reduction of uncertainty (entropy) in semantic maps.

\subsection{Underlying Pose Graph Spectrum for Optimality Criteria}
In addition to the dense mapping uncertainty, another important source of uncertainty comes from the robot's state estimation. It is even more important to analyze the error in pose graph optimization. According to the Cramér-Rao lower bound~\cite{chen2021cramer}, the covariance matrix of any unbiased estimation satisfies the lower bound defined as the Fisher Information Matrix (FIM). The FIM (also noted as Hessian matrix in some literature) is expressed as:
\begin{equation} \label{eq16}
F(x) = \frac{1}{2}\sum\limits_{k = 1}^m {J_k^T(x){{\overline \Sigma  }^{ - 1}}{J_k}(x)},
\end{equation}
where $J$ is the Jacobian matrix of the error cost function with respect to the variable $x$ and ${\overline \Sigma  }^{ - 1}$ is the weighted covariance matrix.
Recent research  \cite{placed2022general} has shown a close relationship between the FIM in the optimization and the Laplacian spectrum of the underlying pose graph structure. This relationship is linked with the different expression using the same FIM:
\begin{equation} \label{eq17}
F(x) = \frac{1}{2}\sum\limits_{k = 1}^m {I_k^T(x)\Sigma _k^{ - 1}{I_k}(x)},
\end{equation}
where 
\begin{equation} \label{eq18}
\Sigma _k^{ - 1} = A{d^T}({T_k})\overline \Sigma  _k^{ - 1}Ad({T_k}),
\end{equation}
where $A{d^T}(T_k)$ is the adjoint matrix of the transformation $T_k$. We recommend readers to see~\cite{placed2022general} for a more detailed derivation. $I_k$ is a sparse matrix consisting of several identity matrices $\mathbb{I}$ filled in $i-$th rows and $k-$th columns. Based on this property, we can rewrite $I_k$ in ${I_k} = a_k^T \otimes\mathbb{I}$ using the Laplacian generator $a$ in section III-C and then equation~\eqref{eq17} turns to:
\begin{equation} \label{eq19}
\begin{split}
F(x) &= \frac{1}{2}\sum\limits_{k = 1}^m {I_k^T(x)\Sigma _k^{ - 1}{I_k}(x)} \\
&\propto \sum\limits_{k = 1}^m {{a_j}a_j^T \otimes \Sigma _k^{ - 1} = \sum\limits_{k = 1}^m {{A_k} \otimes \Sigma _k^{ - 1}} }.
\end{split}
\end{equation}
The estimation uncertainty will keep increasing during exploration or decreasing when a loop closure optimization is performed, leading to a variable covariance matrix. Considering the upper-bound of $\Sigma _k^{ - 1}$, let $(\mu _1^k,\mu _2^k,...,\mu _n^k)$ the eigenvalue of $\Sigma _k^{ - 1}$ in increasing order, it holds $\Sigma _k^{ - 1} \le \mu _n^k\mathbb{I}$. Denote $\Phi_k  = \Sigma _k^{ - 1}$, we will use the FIM $\Phi_k$ to weight every edge in the Laplacian graph. Using the infinite norm property and the concept of the Kronecker product $\otimes$, $\forall k, {\Phi _k} \le {\left\| {{\Phi _k}} \right\|_\infty }$, the following inequality for~\eqref{eq19} holds:
\begin{equation} \label{eq20}
F \le \sum\limits_{k = 1}^m {({{\left\| {{\Phi _k}} \right\|}_\infty }{A_k})}  \otimes\mathbb{I} = {L_\omega} \otimes\mathbb{I},
\end{equation}
where ${L_\omega }$ is the weighted Laplacian graph with ${\omega _k} = {\left\| {{\Phi _k}} \right\|_\infty }$, using Weyl’s monotonicity theorem~\cite{gohberg1978introduction}, \eqref{eq20} turns to:
\begin{equation} \label{eq21}
{\left\| F \right\|_p} \le {\left\| {{L_\omega }} \right\|_p},\;\;\;\omega  = {\left\| \Phi_k  \right\|_\infty },\forall p.
\end{equation}
Equation~\eqref{eq21} gives an upper-bound for ${\left\| F \right\|_p}$. Instead of using the infinite norm of $\Phi_k$, we can relax the inequality relation by weighing every edge of the pose graph using the specific $p$-norm. We will get the following approximation:
\begin{equation} \label{eq22}
{\left\| F \right\|_p} \approx {\left\| {{L_\omega }} \right\|_p},\;\;\;\omega  = {\left\| \Phi_k  \right\|_p},\forall p.
\end{equation}
Equation~\eqref{eq22} gives a general relationship between the pose graph reliability and the underlying Laplacian spectrum, further, using~\eqref{eq8}, the optimality criteria is calculated by:
\begin{equation} \label{eq23}
D \text- opt(F) \approx D \text- opt({L_\omega }) = {(nS({{\cal G}_\omega }))^{\frac{1}{n}}}.
\end{equation}

\subsection{Utility Function}
Consistent with prior work~\cite{7139224,placed2021fast}, we assume that the robot's state and map uncertainties are independent, allowing for the utility function to be calculated as a heuristic linear weighting function. It is remarkable that in some recently advanced SLAM frameworks~\cite{10380742,paloc} the uncertainty of the map points are also treated as a separate model and its influence on state estimation is also considered. Modeling map uncertainty during optimization differs from the active SLAM discussed in this paper. Following the classical exploration pipeline, we only take odometry results from SLAM estimation and we build the map for path planning in a separate dense mapping thread. We define the utility function in the form of Shannon-Rényi entropy as in literature~\cite{7139224}, however, we further improve it in our approach by evaluating all mutual information along each potential path, instead of calculating the proportion of unknown cells only at the frontier. This approach helps to mitigate the stochasticity associated with frontier detection and incorporates the effect of semantic uncertainty and also promotes the selection of trajectories that can improve the overall semantic mapping accuracy, not just at the frontier. The utility function is defined as:
\begin{equation} \label{eq24}
\begin{aligned}
a^* &= \arg \mathop {\max }\limits_a SR(a,T)\\
 &= \arg \max (D \text- opt(L_\omega ^{a,T}) - (\frac{1}{{1 - \alpha }})D \text- opt(L_\omega ^{a,T})),   
\end{aligned}
\end{equation}
where
\begin{equation} \label{eq25}
\alpha  = 1 + \frac{Cost(a,T)}{{\int\limits_{\theta  \in \Theta } {I(m,{{\underline Z }_{t + 1:t + T,\theta }})d\theta } }}
\end{equation}
where $a^*$ is the best action among all potential actions $a$, $SR$ refers to Shannon-rényi entropy, $T$ is the action time, $(D \text- opt(L_\omega ^{a,T})$ and ${\int\limits_{\theta  \in \Theta } {I(m,{{\underline Z }_{t + 1:t + T,\theta }})d\theta } }$ can be calculated by~\eqref{eq23} and~\eqref{eq13}, respectively. Parameter $\alpha$ essentially means using expected information gain to discount the localization error from a trajectory. $Cost(a,T)$ refers to the cost for a path, in this paper, we use traveling distance.
Thanks to the properties of the Shannon-Rényi entropy,~\eqref{eq24} circumvents the drawbacks associated with manually adjusting the weighting parameters in a simple linear weighting function. This approach effectively balances the exploration of new regions—aiming to achieve higher information gains and reduce map uncertainty—with the strategic revisiting of previously explored areas, based on robot localization uncertainty.

Notably, the potential for loop closure allows the robot to reduce its localization uncertainty without revisiting previously explored areas. We take this into account when calculating the D-optimality. Specifically, while weighting the underlying Laplacian graph, those edges likely to form loop closure are assigned greater weight rather than only calculating the FIM. This results in higher D-optimal values for such edges, which implicitly encourages the robot to actively search for loop closures.

\subsection{Autonomous Exploration}
To perform autonomous exploration, we first project the 3D map onto a 2D space. Subsequently, we delineate the boundary between explored and unexplored regions, utilizing edge detection. We then cluster map cells that share connected boundary components, with each cluster representing a frontier. For each frontier, we use the $A^*$ algorithm to generate a path to each center of the frontier. All generated paths will be evaluated using the proposed function. Goals that cannot be reached, either due to the impossibility of path computation or because they were previously identified as frontiers, are blacklisted, indicating that the goal is currently unreachable. From the global path determined by $A^*$, we select several nodes as waypoints for the local planner, lattice space search \cite{zhang2020falco} is used for local planning and obstacle avoidance. Finally, the underlying controller generates the control commands based on the output of the local planner. 
\begin{figure*}[htbp]
    \centering
    \begin{subfigure}[b]{0.23\linewidth}
        \includegraphics[width=\linewidth,height=3.5cm]{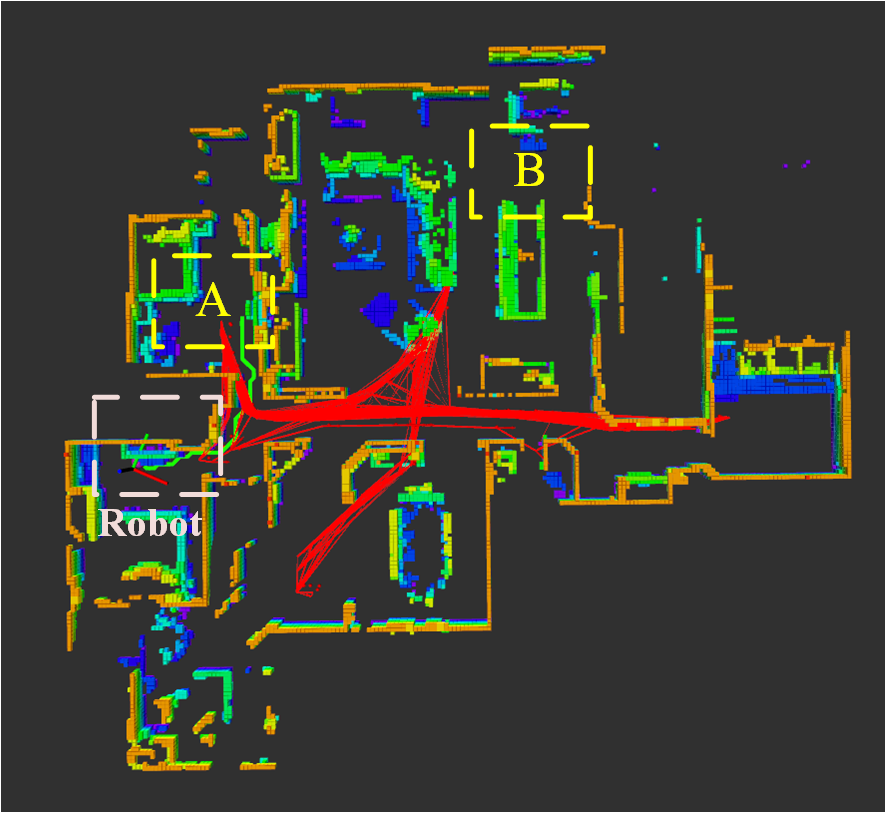}
        \caption{}
        \label{pgo.1}
    \end{subfigure}
    \begin{subfigure}[b]{0.23\linewidth}
        \includegraphics[width=\linewidth,height=3.5cm]{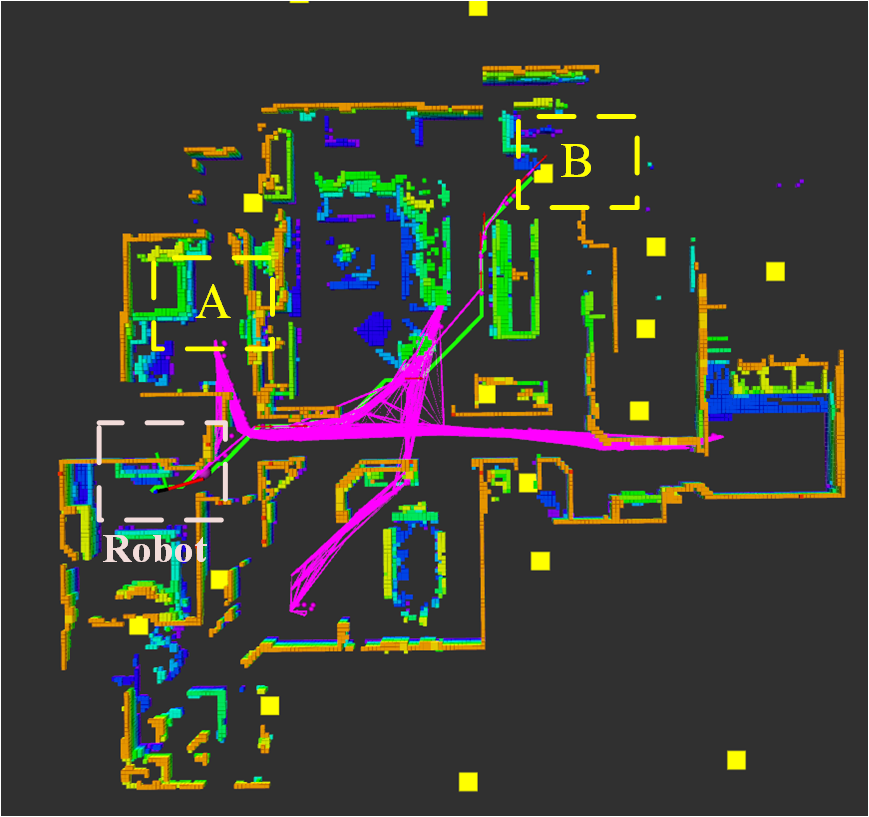}
        \caption{}
        \label{pgo.2}
    \end{subfigure}
    \begin{subfigure}[b]{0.23\linewidth}
        \includegraphics[width=\linewidth,height=3.5cm]{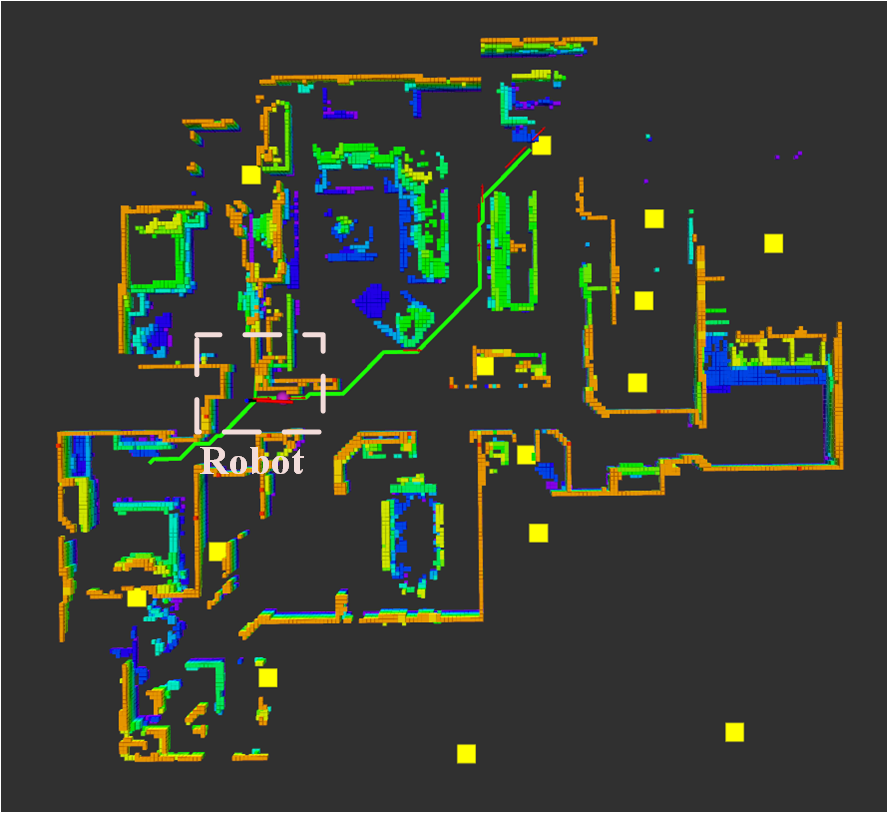}
        \caption{}
        \label{pgo.3}
    \end{subfigure}
    \begin{subfigure}[b]{0.23\linewidth}
        \includegraphics[width=\linewidth,height=3.5cm]{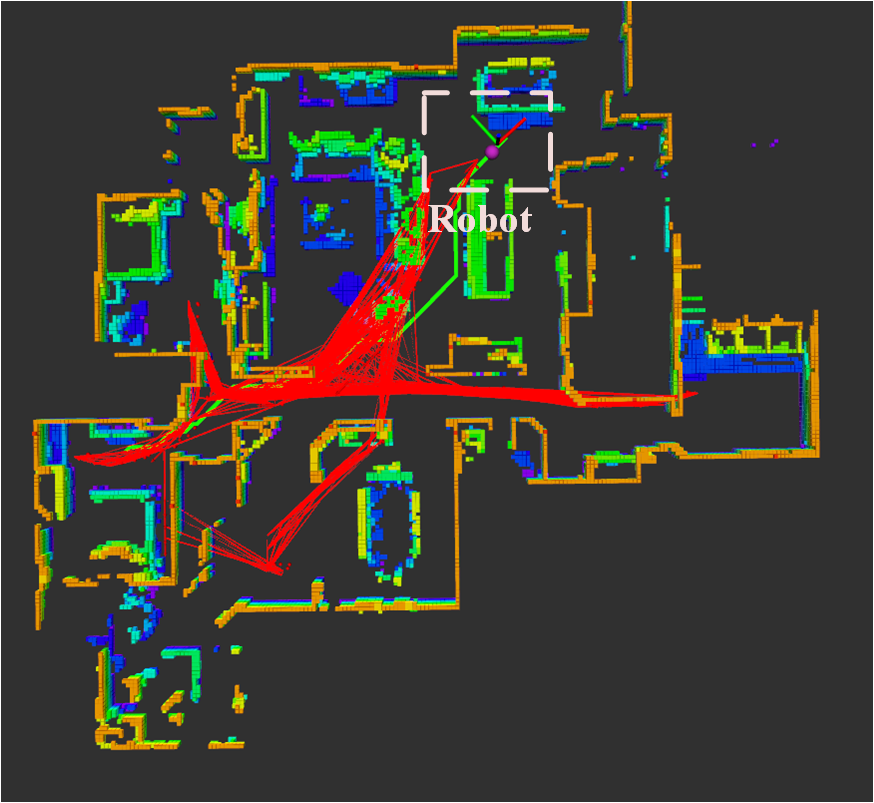}
        \caption{}
        \label{pgo.4}
    \end{subfigure}
    \caption{Schematic diagram of a decision-making process during the active SLAM. The path of frontier A has shorter travel distance while the path of frontier B will provide a more optimistic pose graph with more co-visibility points and loop closure edges for visual SLAM. The final choice of exploration is frontier B. In the figure, red and pink lines represent pose graphs at different stages. Yellow rectangles are candidate frontiers, some of them have been excluded before the decision due to inaccessibility. Green line represents the global path planned by A* algorithm and pink rectangle represents the robot. (a) The robot plans a path toward frontier A and evaluates the corresponding hallucination graph and semantic mutual information. (b) The robot plans a path toward frontier B and evaluates the corresponding hallucination graph and semantic mutual information. (c) The robot executes a path to frontier B. (d) The final generated pose graph after the robot reaches frontier B.}
    \label{pgo}
\end{figure*}

\section{Experiments}
\subsection{Environment Setup}
We evaluate our method in a photorealistic simulation environment~\cite{cmu}. The simulation experiments are conducted on a laptop with an Intel Core i5-11400h CPU and NVIDIA GeForce RTX 3060 GPU. The robot is equipped with an RGB-D camera and the semantic segmentation comes from the Habitat simulator \cite{szot2021habitat}. The semantic segmentation is running at 640$\times$360 resolution with a camera frame rate of 30hz. We adopt the graph-based visual SLAM approach in \cite{7946260}, leveraging the g2o library \cite{kummerle2011g} for backend optimization.  We use a local planner in \cite{cmu} with 3D terrain analysis capabilities. Our environment representation is based on the semantic octomap \cite{hornung13auro}, within which we evaluate semantic mutual information in a 3D space. The map resolution is selected to 0.25 and the semantic mapping is running at 2hz. Although the Habitat simulator is primarily designed for 2D tasks, our approach can be easily extended to 3D environments.

We conducted 10 repetitions of the exploration in two different environments on the Matterport dataset \cite{Matterport3D}, with exploration stopping criteria of 5 minutes for the first smaller building and 10 minutes for the second larger building. We compared our approach with the state-of-the-art methods SSMI \cite{asgharivaskasi2023semantic} and TARE \cite{cao2023representation}, where SSMI is an advanced active semantic mapping method based on information theory, and TARE has been applied in large-scale outdoor environments like the DARPA SubT Challenge. We used the default parameters in their settings. We focus on the localization accuracy and the quality of the semantic maps during exploration, they will be analyzed in section  V-B. In addition, we also compare the exploration efficiency of each method by monitoring the explored area volume over time.

\subsection{Results and Discussion} 
\subsubsection{Geometric SLAM Metric Evaluation}

We employed the widely used  Absolute Trajectory Error (ATE) as the localization evaluation metric and the mean Euclidean distance as the metric for map evaluation as in~\cite{paloc}. 
    
\begin{figure}[htbp]
    \centering
    \begin{subfigure}[b]{0.45\linewidth}
        \includegraphics[width=\linewidth,height=3cm]{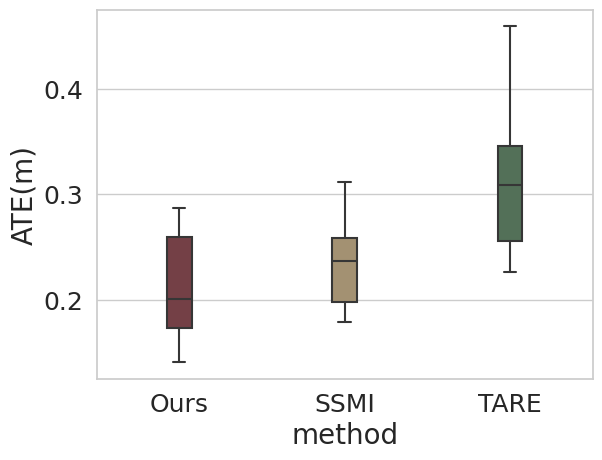}
        \caption{}
        \label{SLAM metric error.1}
    \end{subfigure}
    \quad 
    \begin{subfigure}[b]{0.45\linewidth}
        \includegraphics[width=\linewidth,height=3cm]{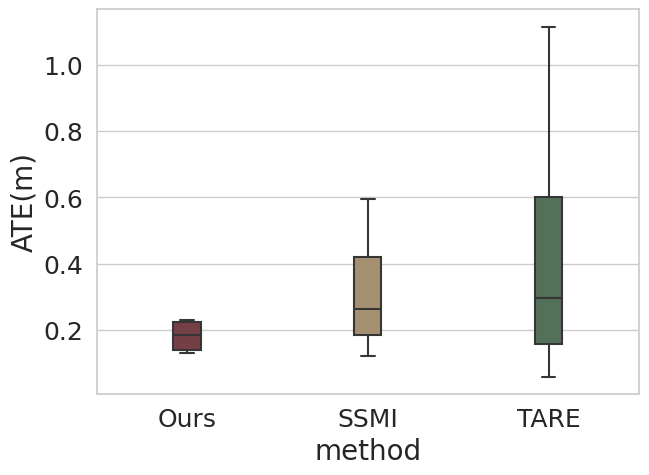}
        \caption{}
        \label{SLAM metric error.2}
    \end{subfigure}

    \begin{subfigure}[b]{0.45\linewidth}
        \includegraphics[width=\linewidth,height=3cm]{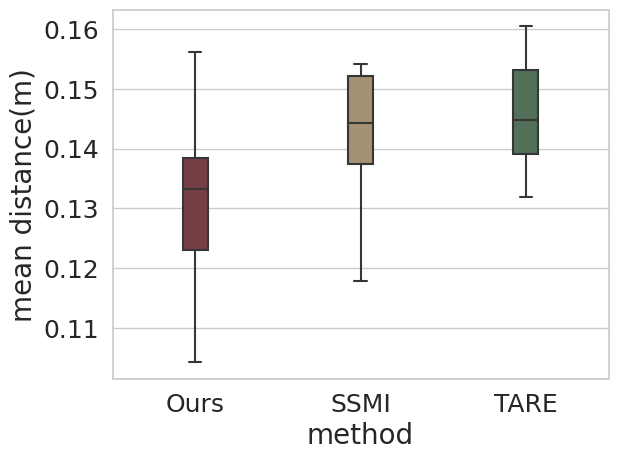}
        \caption{}
        \label{SLAM metric error.3}
    \end{subfigure}
    \quad 
    \begin{subfigure}[b]{0.45\linewidth}
        \includegraphics[width=\linewidth,height=3cm]{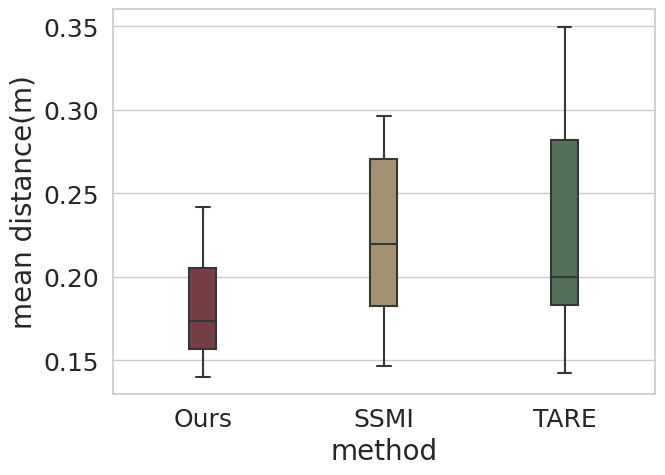}
        \caption{}
        \label{SLAM metric error.4}
    \end{subfigure}
    
    \caption{SLAM error results for 10 repetitions of the exploration experiment. (a) Localization error in experiment 1. (b) Localization error in experiment 2. (c) Map error in experiment 1. (d) Map error in experiment 2.}
    \label{SLAM metric error}
\end{figure}

In practice, we recorded the data from each exploration in the rosbag format. Localization errors are computed using evo~\cite{grupp2017evo}. For semantic mapping, we saved the maps in binary color octomap format, converted into point cloud maps at a resolution of 0.25m for evaluation using the CloudCompare software \cite{cloudcompare}. We plot the distribution of the localization and map errors for each experiment as shown in Fig.~\ref{SLAM metric error}. Compared to the other two methods, our method exhibits a reduction in average localization and map errors with a maximum of 38\% and 21\%. Specifically, our method obtained the smallest localization and map errors in 8 of 10 repeated times in experiment 1, and 9 of 10 in experiment 2. This is attributed to our approach's focus on uncertainty in SLAM pose graph optimization, enabling the robot to intentionally (re)select trajectories with less uncertainty instead of greedily traveling to the place with less traveling distance or information gain. See Fig.~\ref{pgo}.

\begin{figure}[htbp]
    \centering
    \begin{subfigure}[b]{0.45\linewidth}
        \includegraphics[width=\linewidth,height=2.5cm]{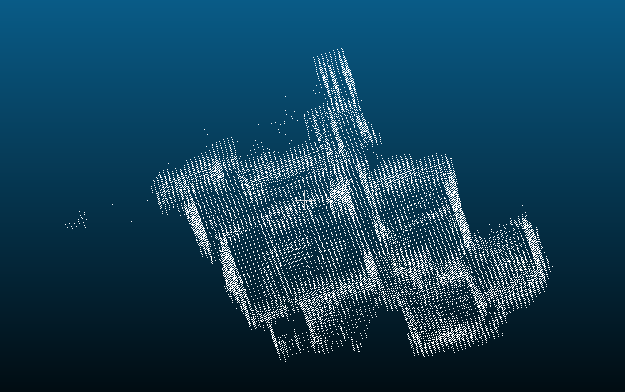}
        \caption{}
        \label{map metric error.1}
    \end{subfigure}
    \quad 
    \begin{subfigure}[b]{0.45\linewidth}
        \includegraphics[width=\linewidth,height=2.5cm]{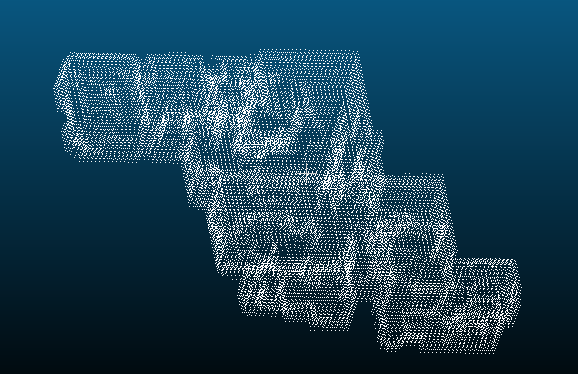}
        \caption{}
        \label{map metric error.2}
    \end{subfigure}

    \begin{subfigure}[b]{0.45\linewidth}
        \includegraphics[width=\linewidth,height=2.5cm]{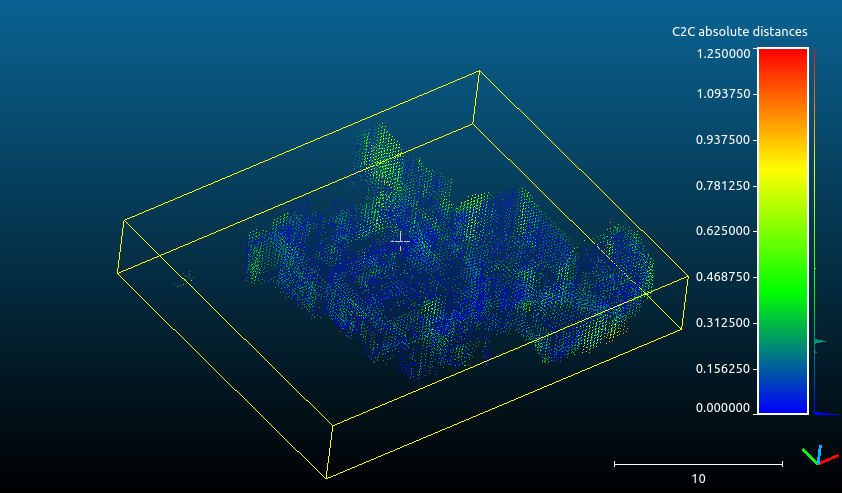}
        \caption{}
        \label{map metric error.3}
    \end{subfigure}
    \quad 
    \begin{subfigure}[b]{0.45\linewidth}
        \includegraphics[width=\linewidth,height=2.5cm]{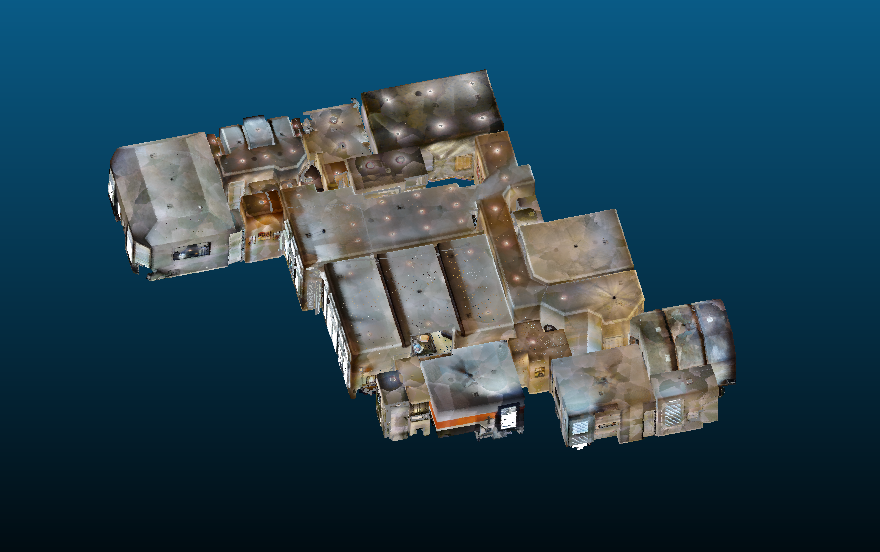}
        \caption{}
        \label{map metric error.4}
    \end{subfigure}

    \begin{subfigure}[b]{0.38\linewidth}
        \includegraphics[width=\linewidth,height=2.0cm]{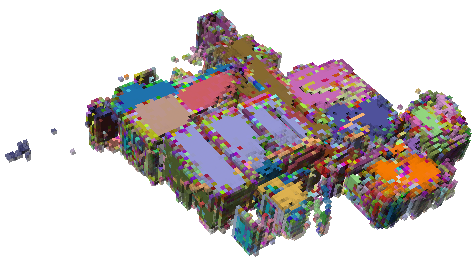}
        \caption{}
        \label{semantic.5}
    \end{subfigure}
    \quad 
    \quad
    \quad
    \begin{subfigure}[b]{0.38\linewidth}
        \includegraphics[width=\linewidth,height=2.0cm]{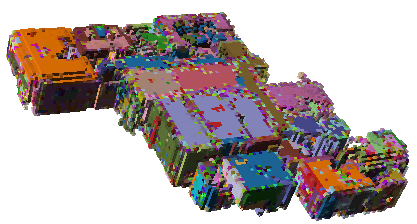}
        \caption{}
        \label{semantic.6}
    \end{subfigure}
    \caption{Visualization of metric-semantic map results obtained during one exploration experiment. (a) Partially explored metric map. (b) Ground truth metric map. (c) The error (Euclidean distance) in CloudCompare software calculation. (d) Ground truth map visualization for the environment. (e) Partially explored semantic map. (f) Ground truth semantic map.}
    \label{map}
\end{figure}

\subsubsection{Semantic Mapping and Classification Results}

As the evaluation method in \cite{gan2019bayesian}, we use Intersection Over Union (IoU) as the semantic map evaluation metric. It is worth noting that the simulator itself does not provide any semantic map, so we initially built a ground truth 3D semantic map based on noise-free state estimation and semantic segmentation from the simulator. Subsequently, we compare the ground truth map with the semantic map constructed upon completion of exploration, as shown in Fig. \ref{map}. The Matterport dataset has 40 instance labels, we focus on the top eight most frequently appearing categories for comparison. The quantitative results are given in Tables~\ref{tab1} and \ref{tab2}.

\begin{table}[htbp] 
\renewcommand\arraystretch{1.0}
    \centering    
    \caption{Quantitative results of experiment 1 for 8 common semantic classes.}
    \label{tab1}    
\resizebox{\columnwidth}{!}{
\begin{tabular}{ccccccccccc}    
\textbf{Metric}&\textbf{Method}&
\cellcolor{objects}\rotatebox{90}{\color{white}objects}&
\cellcolor{table}\rotatebox{90}{\color{white}table}&
\cellcolor{cabinet}\rotatebox{90}{\color{white}cabinet}&
\cellcolor{wall}\rotatebox{90}{\color{white}wall}&
\cellcolor{clothes}\rotatebox{90}{\color{white}clothes}&
\cellcolor{mirror}\rotatebox{90}{\color{white}mirror}&
\cellcolor{picture}\rotatebox{90}{\color{white}picture}&
\cellcolor{lighting}\rotatebox{90}{\color{white}lighting}&
\rotatebox{90}{\textbf{Average}}\\
\midrule
\multirow{3}*{IoU}&Ours&\textbf{47.46}&47.27&\textbf{46.21}&\textbf{45.12}&\textbf{40.87}&\textbf{38.15}&\textbf{37.56}&\textbf{34.56}&\textbf{42.15}\\
&SSMI&45.25&\textbf{48.01}&45.25&42.71&37.55&38.08&36.47&32.42&40.72\\
&TARE&39.62&45.47&39.74&25.35&37.82&23.04&30.34&21.18&32.82\\
\bottomrule
\end{tabular}
}
\end{table}

\begin{table}[htbp] 
\renewcommand\arraystretch{1.0}
    \centering    
    \caption{Quantitative results of experiment 2 for 8 common semantic classes.}
    \label{tab2}    
\resizebox{\columnwidth}{!}{   
\begin{tabular}{ccccccccccc}    
\textbf{Metric}&\textbf{Method}&
\cellcolor{ceiling}\rotatebox{90}{\color{white}ceiling}&
\cellcolor{appliances}\rotatebox{90}{\color{white}appliances}&
\cellcolor{sink}\rotatebox{90}{\color{white}sink}&
\cellcolor{stool}\rotatebox{90}{\color{white}stool}&
\cellcolor{plant}\rotatebox{90}{\color{white}plant}&
\cellcolor{counter}\rotatebox{90}{\color{white}counter}&
\cellcolor{table}\rotatebox{90}{\color{white}table}&
\cellcolor{clothes}\rotatebox{90}{\color{white}clothes}&
\rotatebox{90}{\textbf{Average}}\\
\midrule
\multirow{3}*{IoU}&Ours&\textbf{50.73}&
\textbf{45.26}&
\textbf{43.91}&
43.51&
\textbf{40.42}&
\textbf{39.18}&
\textbf{37.99}&
\textbf{33.57}&
\textbf{41.82}\\
&SSMI&46.02&
41.01&
25.13&
\textbf{43.80}&
39.30&
36.12&
29.25&
28.83&
36.18\\
&TARE&42.01&
36.86&
23.86&
41.17&
32.51&
31.70&
23.27&
30.67&
32.76\\
\bottomrule
\end{tabular}
}
\end{table}
Our method achieves the highest average IoU values across these semantic categories, outperforming other baseline methods with up to a 9.33\% improvement. Our method obtained the highest average IoU value 8 of 10 repeated times in both experiments 1 and 2. The average IoU values of SSMI are higher than those of TARE, as SSMI also takes semantic uncertainty into account. Another interesting observation is that our IoU values match, but are slightly lower than the overall level of multi-class mapping results reported in~\cite{gan2019bayesian}. This discrepancy can be attributed to our utilization of a naive discrete voxel probability fusion updating approach. Integrating the active mapping method proposed in our paper with other, more advanced semantic mapping methods, such as~\cite{gan2019bayesian}, could offer further improvements.

\subsubsection{Exploration Metrics}

\begin{figure}[htbp]
    \centering
    \begin{subfigure}[b]{0.49\linewidth}
        \includegraphics[width=\linewidth]{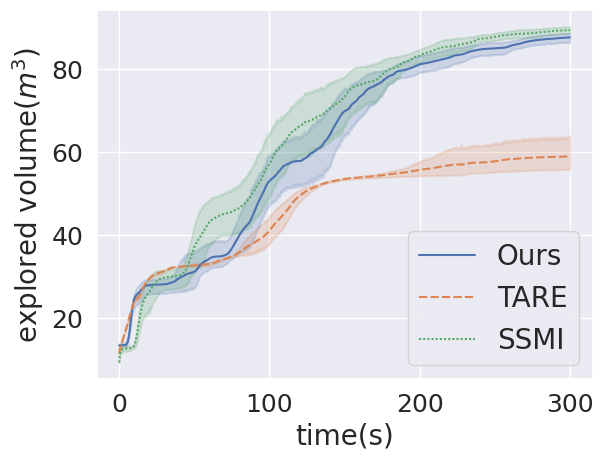}
        \caption{}
        \label{exploration.1}
    \end{subfigure}
    \begin{subfigure}[b]{0.49\linewidth}
        \includegraphics[width=\linewidth]{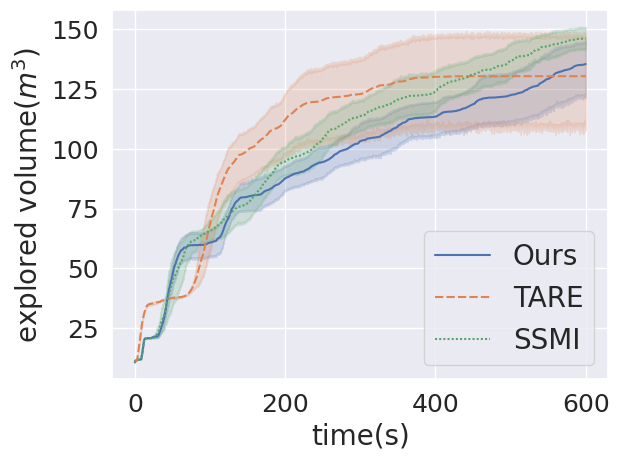}
        \caption{}
        \label{exploration.2}
    \end{subfigure}
    \caption{Comparison of exploration results. (a) The curves of explored volume over time in experiment 1. (b) The curves of explored volume over time in experiment 2.}
    \label{exploration}
\end{figure}

Finally, we investigated the impact of evaluating semantic mutual information as well as pose graph optimality on exploration efficiency. This involved plotting the volume explored over time across multiple iterations of our experiments within two different environments, as illustrated in Fig.~\ref{exploration}. In the plots, the middle line denotes the median value, the upper and lower bounds are represented by the shadow area. Interestingly, we observed partial loss of localization results during some experiments conducted in the baselines, for a fair comparison we excluded these experiments and focused only on those where the entire exploration stack worked well.
The first conclusion from the results is that our method closely matches SSMI in terms of exploration efficiency, albeit with a slight increase in time overhead. This trade-off is considered advantageous, as it aims to enhance SLAM performance and reduce the risk of localization failures, such as tracking lost, which could significantly impact the exploration system. 

The second observation is that the information theory-based exploration strategies (our approach and SSMI) outperform TARE in small-scale environments. However, TARE shows better performance in larger environments due to its hierarchical exploration strategy, which is specifically designed to address large-scale exploration challenges. This work aims to provide a mathematical explanation and complement the state-of-the-art exploration methods from an uncertainty perspective, rather than offering a replacement. 
Using a combination of global and local exploration strategies based on different resolutions rather than information-theoretic methods would be beneficial in accelerating exploration, although this is outside the scope of this paper.

Last, it is also important to mention that as the pose graph grows, the overhead of computing the underlying weighted Laplacian graph may be significant. At this point, spectral measurement sparsification for the pose graph may be a promising tool to further improve exploration efficiency.

\section{Conclusion}
In this paper, we proposed a real-time active metric-semantic SLAM method for efficient and accurate robot exploration. Our method not only evaluates the path with the largest semantic information gain in 3D space, but also integrates the effect of robot state uncertainty. The computation of the large and dense FIM is transformed into the analysis of the underlying pose graph topology, allowing planning to be executed online. The Shannon-Rényi formed entropy avoids different orders of magnitude of uncertainty, leading to an autonomous balance between exploration and exploitation. Comparison with advanced exploration methods including SSMI and TARE demonstrates the performance of our implementation. The research on semantic uncertainty is also meaningful for computer vision tasks, such as mitigating the impact of domain shifts caused by data and scenes. Our implementation can also be used for aerial robot exploration, with a 3D path planner. Future work will focus on spectral graph sparsification to accelerate exploration. We are also interested in generalizing to multi-robot exploration and conducting field experiments.








\bibliography{iros2024}
\bibliographystyle{IEEEtran}

\end{document}